\documentclass[cameraready]{Interspeech}
\usepackage{times}
\usepackage{latexsym}
\usepackage[utf8]{inputenc}
\usepackage{amsmath}
\usepackage{amssymb}
\usepackage{amsfonts}
\usepackage{microtype}
\usepackage[numbers,square]{natbib}
\usepackage{inconsolata}

\usepackage{graphicx}
\usepackage{svg}

\usepackage[T1]{fontenc}

\title{LOPA: Enhancing Spoken Language Assessment via Latent Ordinal Prototype Alignment}

\author[affiliation={1},orcid=0009-0002-7161-7369]{Hong-Yun}{Lin}
\author[affiliation={1}]{Fu-An}{Chao}
\author[affiliation={1}]{Bi-Cheng}{Yan}
\author[affiliation={1}]{Berlin}{Chen}

\address{ $^1$ National Taiwan Normal University }

\email{buffett@ntnu.edu.tw, fuanchao@ntnu.edu.tw, 80847001s@ntnu.edu.tw, berlin@ntnu.edu.tw}


\usepackage{comment}

\begin{document}

\maketitle

\begin{abstract}
Fueled by increasing model scale and multimodal inputs, Multimodal Large Language Models (MLLMs) have emerged as a promising paradigm for Spoken Language Assessment (SLA). While effective, this paradigm often overlooks the intrinsic ordinal structure of language acquisition. This paper works around the necessity of large-scale MLLMs by introducing Latent Ordinal Prototype Alignment (LOPA) for SLA, a prototype-based regularizer that enforces an ordinal geometric prior directly on the latent space. Coupled with Semantic-Anchored Layer Routing (SALR), which adaptively harvests multi-depth representations from a frozen Whisper encoder, our framework achieves an RMSE of 0.361. This performance rivals billion-parameter systems without the need for LLM-based fine-tuning. Further analysis reveals that SALR's synergy with LOPA offers interpretable, criterion-aligned preferences, thereby supporting an efficient and ordinal-aware modeling alternative to current scaling-centric models for SLA.
\end{abstract}

\section{Introduction}

Spoken Language Assessment (SLA) aims to estimate a learner's oral proficiency from spontaneous
spoken responses, providing crucial support for Computer Assisted Language Learning (CALL)
\cite{IEEE5881478}. Most recently, the emergence of large-scale multimodal Large Language Models
(MLLMs) has significantly reshaped the landscape of SLA. To harness the power of these models, prior
efforts typically adopt supervised fine-tuning (SFT) and instruction tuning \cite{ma25b_interspeech,
lin2025sessionlevelspokenlanguageassessment} to achieve impressive performance on SLA. However, this
paradigm inevitably faces two critical challenges. First, the computational cost of fine-tuning and
deploying billion-parameter models would be prohibitive for many real-world educational use cases
under low-resource settings. Second, these models often treat proficiency scoring as a
text-generation or standard regression problem, which largely ignores the inherent ordinal structure
that reflects the progressive nature of human language acquisition.

Emerging as a practical alternative, recent research has pivoted toward leveraging moderate-sized
speech foundation models like Whisper \cite{radford2023robust}. While these approaches are more
lightweight, they typically rely on the final encoder layer \cite{chao2026probinghiddentalentasr} or
transcription-based features \cite{cai25_slate}, thereby discarding the rich acoustic, phonetic and
phonological cues preserved in intermediate representations. Furthermore, existing Whisper-based
systems generally omit ordinal information, treating proficiency levels as independent categories
rather than a continuous, developmental trajectory. This limitation is particularly detrimental to
SLA, where the distinction between adjacent levels (e.g., B1 vs. B2) requires a nuanced
understanding of both semantic depth and articulate precision.

In light of these challenges, we in this paper advocate that principled architectural design can be
more effective than straightforward scaling for language assessment. We propose a lightweight
framework that exploits multi-layer representations from a frozen Whisper encoder through SALR. This
allows the model to dynamically integrate information across different abstraction levels, from deep
semantic cues to shallow acoustic features. In addition, to better capture the progressive nature of
language proficiency, we introduce Latent Ordinal Prototype Alignment (LOPA), which effectively
integrates an ordinal prototype loss that enforces ordinal relationships directly in the embedding
space for SLA. Without fine-tuning the backbone or employing an LLM, our approach achieves
performance comparable to substantially larger multimodal systems, while featuring simplicity and
explainable behavior, which makes the analysis easier.

\begin{figure*}[t]
    \centering
    \includegraphics[width=0.75\textwidth]{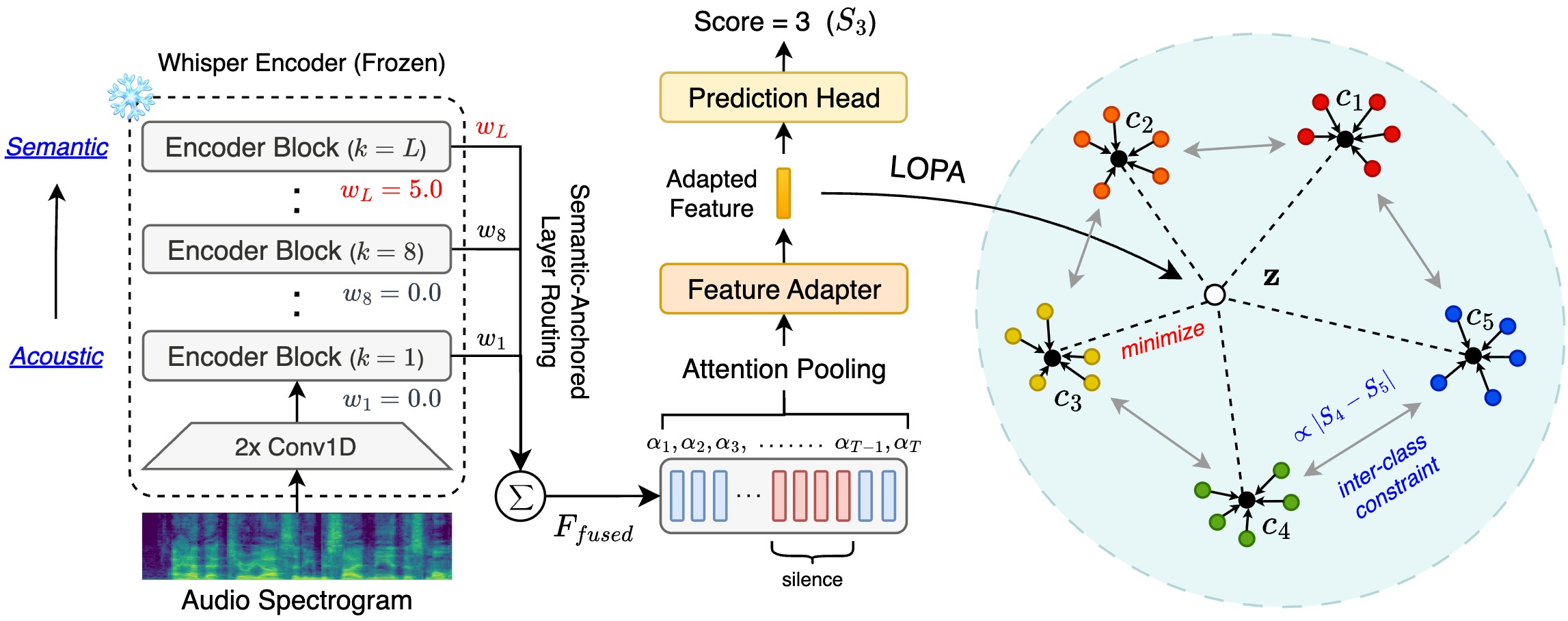}
    \caption{A schematic diagram of the proposed method.}
    \label{fig:overview}
\end{figure*}

\section{Related Work}
\label{sec:motivation}
Recent studies have noted that large-scale pre-trained speech models, such as Whisper
\cite{radford2023robust}, encode information in a strictly hierarchical manner. Specifically,
\cite{klimova2024uncovering} indicated that syllabic and phonetic cues are concentrated in the
lower-to-middle encoder layers. This aligns with findings in \cite{gandelsman2024beyond}, which
demonstrated a transition from low-level acoustic dominance in early layers to linearly separable
semantic attributes in the higher layers. Building on this, \cite{chao2025probinghiddentalentasr}
pointed out that Whisper’s final encoder layer can be effectively probed for SLA, revealing an
implicit ordinal structure aligned with CEFR proficiency levels \cite{council2001common}. However,
focusing solely on the final layer risks discarding fine-grained acoustic/phonetic information
crucial for evaluating pronunciation and fluency, and the resulting latent space remains weakly
separated, with substantial overlap between proficiency levels.

As a remedy, prototype-based formulations provide a simple mechanism to encourage class-wise
compactness in embedding spaces and have been widely used as a geometric inductive bias in
representation learning \cite{NIPS2017_cb8da676,ECCV2016_10.1007}. More broadly, SUPERB
\cite{yang21c_interspeech,shi23g_interspeech} adopts a frozen-backbone, lightweight-head framework
and explicitly aggregates multiple hidden states via a weighted sum, revealing that the last-layer
representation is not always optimal across downstream speech tasks.

Collectively, these findings suggest that single-layer probing provides an incomplete
characterization of learner speech, motivating our approach that integrates multi-layer
representations to capture both low-level speech attributes and high-level proficiency signals while
explicitly enhancing ordinal separability across proficiency levels.

\section{Methodology}
\label{sec:methodology}

\subsection{Problem Formulation}
This paper focuses on the task of spoken language assessment (SLA) in a multi-part test setting. A
test-taker provides spoken responses across different parts $\mathcal{P} = \{P1, P3, P4, P5\}$,
where each part is designed to elicit specific language competencies. The goal is to learn a
regression function $f: \mathcal{X} \to y$ that maps the speech input $\mathcal{X}$ of a specific
part to its corresponding proficiency score $y$. Reference scores are provided by human graders on
an ordinal proficiency scale (\ref{sec:dataset}); we aim to train auto-graders whose predictions
closely match these references.

\subsection{Model Architecture}
The proposed method transforms an input speech waveform $\mathcal{X}$ into a proficiency score
$\hat{y}$ through a four-stage pipeline designed to maximize the extraction of important and
relevant information while maintaining the interpretability of prediction results. An overview of
the complete workflow of our proposed method is depicted in Figure~\ref{fig:overview}.

\textbf{Stage 1: Multi-Layer Feature Extraction}
We utilize the encoder of a pre-trained speech foundation model as a frozen backbone. Given an input
audio spectrogram, the encoder outputs a stack of hidden states $H = \{H_1, H_2, \dots, H_{L}\}$,
where $H_l \in \mathbb{R}^{T \times D}$ represents the sequence of $T$ token embeddings from the
$l$-th layer, and $D$ is the feature dimension. 

\textbf{Stage 2: Semantic-Anchored Layer Routing (SALR)}
To synthesize these multi-level features, we use a lightweight layer-weighting module named
Semantic-Anchored Layer Routing to form a semantically anchored mixture of encoder layers. A set of
scalar weights $\{w_l\}_{l=1}^{L}$ is learned to compute a weighted sum of the layers:
\begin{equation}
    F_{\text{fused}} = \sum_{l=1}^{L} \frac{\exp(w_l)}{\sum_{k=1}^{L} \exp(w_k)} H_l.
\end{equation}
Motivated by the acoustic-to-semantic stratification observed in Whisper layers
(Section~\ref{sec:motivation}), the routed mixture can selectively incorporate complementary depths
for proficiency scoring. 
To stabilize training, we initialize $w_L$ to a high value and the others to zero. This biased
initialization anchors training on a strong semantic baseline and allows the model to recruit
additional layers only when beneficial, thereby mitigating feature dilution from indiscriminate
mixing in our frozen-backbone setting.

\textbf{Stage 3: Attention-based Temporal Pooling}
Since spoken responses often vary in length and saliency, we apply attention pooling to aggregate
$F_{\text{fused}}$ into a fixed-size vector. We compute a scalar score $u_t$ for each time step
using a shared scorer and normalize it with a masked softmax, yielding attention weights $\alpha_t$
with $\sum_{t=1}^{T}\alpha_t=1$. The pooled representation is expressed by
\begin{equation}
\mathbf{h}_{\text{pool}} = \sum_{t=1}^{T} \alpha_t F_{\text{fused}, t}.
\end{equation}

\begin{figure*}[t]
    \centering
    \includegraphics[width=0.72\textwidth]{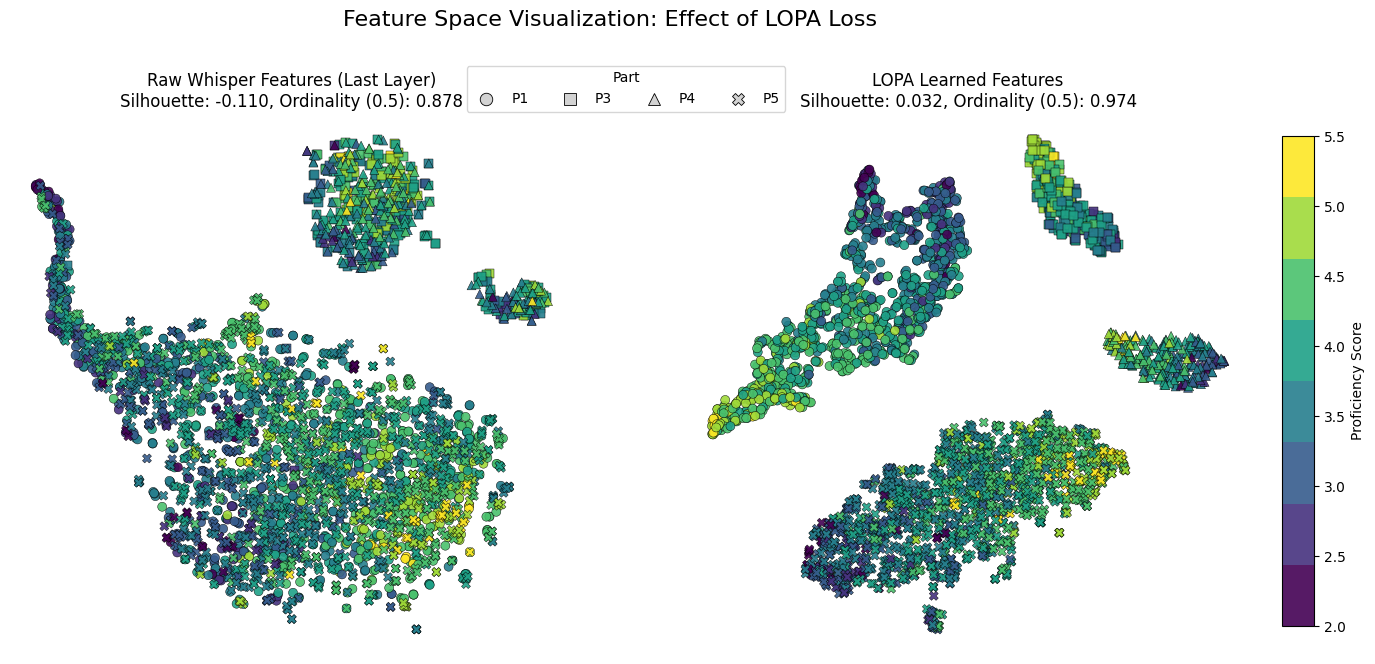}
    \caption{t-SNE visualization of the latent space. \textbf{Left:} raw last-layer Whisper
    representations exhibit noticeable overlap between adjacent proficiency levels. \textbf{Right:}
    with our LOPA loss, embeddings become more compact within levels and more separable across
    levels, forming a clearer ordinal trajectory consistent with CEFR progression. Marker types
    denote the four test parts (P1/P3/P4/P5).}
    \label{fig:tsne_lopa}
\end{figure*}

\begin{table*}[t]
\centering
\small
\setlength{\tabcolsep}{6pt}
\caption{ Overall performance on the S\&I evaluation set. Our Whisper-only model achieves
competitive accuracy against multimodal graders without multimodal LM fine-tuning. }
\begin{tabular}{l|l|c|c|c|c}
\hline
\textbf{Family} & \textbf{Model} & \textbf{RMSE} & \textbf{PCC} & \textbf{\%$\le$0.5} &
\textbf{\%$\le$1.0} \\
\hline
Lightweight & BERT (cascaded ASR $\rightarrow$ BERT baseline) \cite{lin25_slate} & 0.445 & 0.727 &
76.0 & 96.3 \\
Lightweight & W2V (wav2vec2 end-to-end grader) \cite{lin25_slate} & 0.394 & 0.790 & 81.3 & 99.3 \\
Lightweight & APP (Whisper last-layer) \cite{lin2025sessionlevelspokenlanguageassessment} & 0.383 &
0.805 & 81.7 & 99.0 \\
Lightweight & Perezoso (Whisper + BERT + handcrafted features) \cite{cai25_slate} & 0.364 & 0.826 &
83.0 & \textbf{99.7} \\
Lightweight & \textbf{Ours (Whisper-only + SALR + LOPA)} & 0.361 & \textbf{0.828} & 83.3 & 99.0 \\
\hline
MLLM & Phi-4-CTG \cite{lin2025sessionlevelspokenlanguageassessment} & 0.412 & 0.796 & 74.7 & 98.0 \\
MLLM & Phi-4-STG \cite{lin2025sessionlevelspokenlanguageassessment} & 0.375 & 0.820 & 81.7 & 99.3 \\
MLLM & Phi-4-MTL \cite{lin2025sessionlevelspokenlanguageassessment} & 0.362 & 0.825 & 85.7 & 99.0 \\
MLLM & Phi-4-MTL-APP \cite{lin2025sessionlevelspokenlanguageassessment} & \textbf{0.360} & 0.827 &
\textbf{85.7} & 99.0 \\
\hline
\end{tabular}
\label{tab:overall_main}
\end{table*}

\textbf{Stage 4: Latent Projection and Scoring}
The pooled representation $\mathbf{h}_{\text{pool}}$  is subsequently transformed via a feature
adapter into a structured latent embedding $\mathbf{z} \in \mathbb{R}^{d_{\text{latent}}}$. In turn,
a linear projection head maps $\mathbf{z}$ to $K$ class logits, which are softmax-normalized to
produce the probability distribution $P(k \mid \mathbf{z})$. To enable continuous scoring and
alleviate information loss in discrete categorical classification \cite{ma25b_interspeech}, the
final proficiency score $\hat{y}$ is computed as the expectation over the ordered score set
$\{s_k\}_{k=1}^{K}$:
\begin{equation}
    {\hat{y} = \sum_{k=1}^{K} P(k \mid \mathbf{z}) \cdot s_k}.
\end{equation}
The model is optimized using $\mathcal{L}_{\text{task}} = \left\lVert \hat{y} - y \right\rVert_2^2$,
where $y$ denotes the ground-truth proficiency score.

\subsection{Latent Ordinal Prototype Alignment (LOPA)}
While raw Whisper representations exhibit an inherent ordinal structure
\cite{chao2025probinghiddentalentasr}, the latent space lacks the distinct separability required for
precise scoring. To sharpen this implicit manifold, we introduce Latent Ordinal Prototype Alignment
(LOPA). Rather than building a structure from scratch, LOPA acts as a geometric regularizer that
amplifies ordinality by aligning embeddings $\mathbf{z}$ with learnable level prototypes
$\mathcal{C} = \{\mathbf{c}_1, \dots, \mathbf{c}_K\}$. Each $\mathbf{c}_k$ is initialized from the
$k$-th CEFR level centroid and updated during training. Let $k_i \in \{1, \dots, K\}$ be the ordinal
index for score $s_{k_i}$. The objective comprises two primary terms:

\subsubsection{Prototype Attraction Loss}
This term induces intra-class compactness by minimizing the Euclidean distance between a sample
representation $\mathbf{z}_i$ and its corresponding ground-truth prototype $\mathbf{c}_{k_i}$:

\begin{equation}
\mathcal{L}_{\text{attract}}
= \frac{1}{N}\sum_{i=1}^{N} \left\lVert \mathbf{z}_i - \mathbf{c}_{k_i} \right\rVert_2^2.
\end{equation}

\subsubsection{Ordinal Constraint Loss}
To impose inter-class topology, we require the geometric distance between prototypes to mirror the
semantic distance between their scores. Let $s_j$ denote the numerical score of class $j$. We learn
a global scaling factor $\alpha>0$ to align the magnitude of prototype distances with score gaps.
The ordinal loss is defined by

\begin{equation}
\mathcal{L}_{\text{ordinal}}
= \frac{1}{K^2}\sum_{j=1}^{K}\sum_{k=1}^{K}
\left( \alpha \left\lVert \mathbf{c}_j - \mathbf{c}_k \right\rVert_2 - \left| s_j - s_k \right| \right)^2.
\end{equation}
This serves as a geometric regularizer that enforces monotonicity across the prototypes, ensuring
that the inter-prototype distances are proportional to the actual proficiency score gaps. The total
objective is:
\begin{equation}
\mathcal{L}_{\text{total}}
= \mathcal{L}_{\text{task}}
+ \lambda_{\text{att}} \mathcal{L}_{\text{attract}}
+ \lambda_{\text{ord}} \mathcal{L}_{\text{ordinal}}.
\end{equation}

While prototypical learning \cite{NIPS2017_cb8da676} and ordinal ranking \cite{gong2022ranksim} have
been explored in isolation on some machine-learning applications, their systematic integration
remains underexplored for speech-related tasks like SLA. LOPA addresses this by fusing these two
paradigms into a unified objective, explicitly shaping the latent space into an ordered manifold.
When coupled with SALR, this lightweight framework provides a novel, task-aware instantiation that
transforms multi-depth representations from a frozen Whisper encoder into structured, actionable
features. The synergy of LOPA and SALR not only enables high-accuracy automated scoring but also
facilitates fine-grained, interpretable analysis of layer-wise model behavior across assessment
criteria.


\section{Experiments}
\label{sec:experiments}

\subsection{Dataset and Protocol}
\label{sec:dataset}
This work evaluates on the Speak \& Improve (S\&I) Corpus 2025 \cite{qian25_slate,knill25_slate}. We
use the official Train/Dev/Eval splits. Following the official SLA setting, we focus on the open
speaking parts $\{P1,P3,P4,P5\}$, with part-level reference scores assigned by human graders on a
CEFR-aligned ordinal scale with half-point increments (2.0--5.5). The overall session score is
defined as the arithmetic mean of the four part scores.

\subsection{Experimental Setup and Compared Approaches}
\label{sec:exp_setup}

Following Sec.~\ref{sec:dataset}, we train part-specific graders for each part to account for
systematic length differences across parts. We compare our method, which utilizes a frozen Whisper
encoder with SALR and a LOPA-trained Feature Adapter, against two baseline categories: \textbf{(i)
Encoder-based or cascaded baselines}: Includes wav2vec2 \cite{baevski2020wav2vec} and
ASR$\rightarrow$BERT graders \cite{devlin2019bert}, a handcrafted feature-augmented Whisper/BERT
cascaded grader (Team Perezoso) \cite{cai25_slate}, and a Whisper final-layer MLP grader (APP)
\cite{lin2025sessionlevelspokenlanguageassessment}. \textbf{(ii) Multimodal foundation-model
baselines}: Fine-tuned Multimodal LLMs for SLA, specifically Phi-4-multimodal variants (STG/CTG and
multi-target learning) from \cite{lin2025sessionlevelspokenlanguageassessment}.

\begin{table}[t]
\centering
\small
\setlength{\tabcolsep}{4pt}
\caption{Ablation study on the evaluation set.}
\begin{tabular}{l|c|c}
\hline
\textbf{Setting} & \textbf{RMSE} & \textbf{PCC} \\
\hline
{\bfseries Full (LOPA + SALR + temporal attn)} & \textbf{0.3618} & \textbf{0.8276} \\
w/o temporal attn (mean over time) & 0.3698 & 0.8220 \\
w/o SALR (only the last layer) & 0.3739 & 0.8192 \\
w/o LOPA (pooling unchanged) & 0.3831 & 0.8052 \\
naive learned layers + mean time & 0.4945 & 0.6897 \\
\hline
\end{tabular}
\label{tab:ablation}
\end{table}

\begin{table}[t]
\centering
\small
\setlength{\tabcolsep}{4pt}
\caption{Significance test on the S\&I evaluation set: paired t-test on examinee-level overall squared errors for Full vs.\ w/o LOPA.}
\begin{tabular}{l|c}
\hline
\textbf{Statistics} & \textbf{Value} \\
\hline
$t$ (df=299) & 2.4345 \\
$p$ (two-sided) & 0.0155 \\
$p$ (one-sided; LOPA better) & 0.0078 \\
Cohen's $d_z$ & 0.1406 \\
\hline
\end{tabular}
\label{tab:significance_lopa}
\end{table}

\subsection{Implementation Details}
\label{sec:impl}
We use the Whisper Large-v3 encoder as a frozen backbone and extract all $L=32$ layer
representations ($D=1280$) \cite{radford2023robust}. SALR is bias-initialized to favor the final
layer (weight=5.0) while others start at 0.0. Temporal aggregation uses attention pooling (hidden
size 128). The pooled vector is passed through a two-layer MLP feature adapter (512 hidden units,
GELU, dropout 0.1) to produce the latent embedding optimized by LOPA. We train with AdamW (lr
$1\times10^{-3}$, batch size 32) for 25 epochs (P1/P5) and 30 epochs (P3/P4), and set
$\lambda_{\text{att}}=\lambda_{\text{ord}}=0.1$.

\subsection{Evaluation Metrics}
\label{sec:metrics}

We report RMSE, PCC, and tolerance accuracies within $\pm0.5$ and $\pm1.0$ on the S\&I scale, and
measure ordinal structure using Silhouette Score \cite{rousseeuw1987silhouettes} and Ordinality
Correlation (OC). OC computes score-wise centroids $\mu_s$ in the embedding space and measures the
Pearson correlation between pairwise centroid distances and the corresponding score gaps:
\begin{equation}
\mathrm{OC}=\mathrm{corr}\Big(\{\lVert \mu_s-\mu_t\rVert_2\}_{s<t},\{|s-t|\}_{s<t}\Big).
\end{equation}

\section{Results and Discussion}
\label{sec:results}

\subsection{Main Results on the S\&I Evaluation Set}
\label{sec:main_results}

Table~\ref{tab:overall_main} summarizes the overall performance on the S\&I evaluation set. With a
frozen Whisper backbone and a lightweight head only, our approach achieves an RMSE of 0.361 and a
PCC of 0.828. This performance is competitive with strong multimodal foundation-model approaches
while avoiding multimodal LLM fine-tuning. Compared to the Whisper last-layer baseline (APP), the
improvements support two key design choices: (i) \textbf{SALR} leverages intermediate
representations to recover acoustic/phonetic cues that are attenuated in the final layer, and (ii)
\textbf{LOPA} explicitly strengthens ordinal structure in the latent space, which is essential for
proficiency scales with ordered levels.

Beyond overall accuracy, the learned SALR weights provide an interpretable signal of part-dependent
feature usage: while the final layer serves as a semantic anchor, auxiliary layers shift across
parts, supporting the notion that different sub-tasks rely on cues at different depths.

\subsection{Does LOPA Improve Ordinal Structure and Reliability?}
\label{sec:lopa_analysis}

Fig~\ref{fig:tsne_lopa} visualizes the latent space before and after applying LOPA. Raw Whisper
last-layer representations exhibit partial clustering but still show substantial overlap between
adjacent proficiency levels. Across all four parts, LOPA yields a more coherent score-ordered
geometry: the global ordinality score increases from 0.878 to 0.974, and the Silhouette score
improves from $-0.110$ to 0.032. Qualitatively, each part exhibits a clearer progression along the
proficiency gradient with reduced overlap between adjacent score bands, suggesting that the
strengthened ordinal structure is consistent rather than being driven by a single part.

Beyond visualization, we assess whether the performance gain from LOPA is reliable rather than
incidental. Using examinee-level overall squared errors (SE), we conduct a paired t-test between
\emph{Full} and \emph{w/o LOPA}. Table~\ref{tab:significance_lopa} shows a statistically significant
error reduction ($p<0.02$ two-sided), supporting that LOPA yields a consistent improvement even when
headline metrics are already in a near-SOTA regime.

\subsection{Layer Preference and Interpretability}
\label{sec:layer_pref}

To understand how SALR utilizes Whisper representations, we inspect the learned layer weights and
report the strongest layers selected for each part in Table~\ref{tab:layer_pref}. While the routing
distribution is typically concentrated on the top layer, the most salient \emph{auxiliary} layers
vary systematically across parts: P1 favors shallow layers, P3/P4 tend to draw from mid-depth
layers, and P5 relies more on higher layers. This part-dependent pattern is consistent with the
intuition that different SLA sub-tasks emphasize cues at different abstraction levels (e.g.,
lower-level acoustic/phonetic cues vs.\ higher-level semantic content), and provides an
interpretable signal of how multi-depth information is leveraged in our framework.

\begin{table}[t]
\centering
\small
\setlength{\tabcolsep}{4pt}
\caption{SALR layer preferences by part. }
\begin{tabular}{l|l}
\hline
\textbf{Part} & \textbf{Top 3 layers (index: weight)} \\
\hline
P1 (Personal QA)  & L32: 0.8489, L02: 0.0054, L01: 0.0053 \\
P3 (Opinion)  & L32: 0.8837, L12: 0.0040, L14: 0.0040 \\
P4 (Graphic)  & L32: 0.8681, L14: 0.0043, L09: 0.0043 \\
P5 (Topic Talk)  & L32: 0.8889, L31: 0.0039, L27: 0.0038 \\
\hline
\end{tabular}
\label{tab:layer_pref}
\end{table}


\subsection{Ablation Study}
\label{sec:ablation}
Table~\ref{tab:ablation} reports an ablation on the evaluation set. Removing LOPA yields a clear
performance drop, supporting the role of ordinal regularization, and removing SALR underperforms the
full model, indicating the benefit of selective multi-depth feature utilization. The naive
layer-aggregation baseline trains 32 uniformly initialized learnable layer weights and uses mean
temporal pooling; its inferior result highlights the need for SALR's bias-initialized semantic
anchor.

\section{Conclusion}
This paper introduced a lightweight Whisper-only SLA framework without backbone fine-tuning.
\textbf{SALR} is a bias-initialized layer router that keeps a semantic anchor while selectively
incorporating cues from different Whisper depths. \textbf{LOPA} explicitly shapes the latent space
with ordinal geometry to match proficiency scales. On Speak \& Improve 2025, our approach reaches
\textbf{RMSE 0.361} with a frozen backbone and lightweight head, and SALR weights provide
part-dependent interpretability.

\section{Use of Generative AI Disclosure}
The authors utilized ChatGPT 5.2 to refine the wording clarity of this manuscript (viz, grammatical
error check and writing style polishing). The authors maintained full control over the research
content and remain responsible for the accuracy and integrity of the experimental results and the
presented figures.

\section{Acknowledgements}
This work was supported by the Language Training and Testing Center, Taiwan. Any findings and
implications in the paper do not necessarily reflect those of the sponsor.

\section{References}
\bibliographystyle{IEEEtran}
\begingroup
\renewcommand{\section}[2]{}
\bibliography{mybib}
\endgroup

\end{document}